# Lessons from the Cruise Robotaxi Pedestrian Dragging Mishap

**Philip Koopman**
Carnegie Mellon University

■ **A robotaxi dragged a pedestrian** 20 feet down a San Francisco street on the evening of October 2, 2023, coming to rest with its rear wheel on that woman's legs. The mishap was complex, involving a first impact by a different, human-driven vehicle. The following weeks saw Cruise stand down its road operations amid allegations of withholding crucial mishap information from regulators. The pedestrian has survived her severe injuries, but the robotaxi industry is still wrestling with the aftermath.

Key observations include that the robotaxi had multiple possible ways available to avoid initial impact with the pedestrian. Limitations to the computer driver's programming prevented it from recognizing a pedestrian was about to be hit in an adjacent lane, caused the robotaxi to lose tracking of and then in essence forget a pedestrian who was hit by an adjacent vehicle, and forget that the robotaxi had just run over a presumed pedestrian when beginning a subsequent repositioning maneuver. The computer driver was unable to detect the pedestrian being dragged even though her legs were partially in view of a robotaxi camera. Moreover, more conservative operational approaches could have avoided the dragging portion of the mishap entirely, such as waiting for remote confirmation before moving after a crash with a pedestrian, or operating the still-developing robotaxi technology with an in-vehicle safety driver rather than prioritizing driver-out deployment.

This paper summarizes the events of the mishap itself, immediate responses, and longer-term regulatory interactions. The emphasis is on the main lessons for autonomous vehicle stakeholders that might be learned. The information is based on an extensive (but partially redacted, and in some ways incomplete) external analysis commissioned by Cruise [1]. Factual statements, quoted text, subjective characterizations of events, attributed attitudes, motive, intent, and other descriptions regarding the mishap and involved parties are sourced from that external analysis unless otherwise indicated. Characterizations of things as having gone right vs. wrong and as lessons that might be learned are those of the author. This paper is a simplified description that draws upon a more detailed analysis with fine-grain source material references [2].

**Before the mishap**

In 2023, San Francisco was a high-profile hotbed of robotaxi deployments. Most notably, hundreds of robotaxis from Cruise and Waymo were providing a combination of crewed and uncrewed taxi service within the city.

Cruise robotaxis were remotely monitored by the Cruise Remote Assistance Center (CRAC) located in Arizona, several hundred miles away. Remote interventions only occurred in response to requests initiated by the autonomous robotaxi's computer driver. CRAC was not tasked with remotely driving vehicles (remote steering wheels were not involved in the operational concept), nor did it exercise continuous driving behavior safety oversight. The robotaxis were on their own for making driving decisions moment-by-moment, and for deciding if and when to pause to wait for remote support guidance.

Numerous complaints about robotaxis causing road disruptions had been accumulating for months, including dozens of complaints by city emergency responders of blocked roads and response scene disruption [3]. After a contentious public hearing, the California Public Utilities Commission (CPUC) granted expanded operational permission for both Cruise and Waymo robotaxi operations to charge fares to the general public. Cruise proceeded to scale up operations.

On August 17, 2023, a week after receiving CPUC authorization to charge fares for passenger service, a Cruise robotaxi collided with a fire truck. The fire truck had its emergency annunciators active (lights, siren, horn). The robotaxi was proceeding through a green light, but failed to yield to the cross-traffic fire truck as required by road rules. As a result, the California Department of Motor Vehicles (DMV) restricted the number of Cruise robotaxis that could be operational at any given time, effectively putting them on a sort of probation [4].

Cruise continued its strategy of scaling up deployment within the constraints of the DMV limitation. This included plans to launch operations in Los Angeles and other cities.

**The initial collisions**

At 9:29 PM on October 2, 2023, an unoccupied Cruise robotaxi operating with no human driver oversight was stopped at a red light in the curb lane of a 4-way intersection of two city surface streets. Next to the robotaxi, in the inner lane of two lanes in that direction, was a human-driven Nissan. When the traffic light changed green, both vehicles accelerated into the intersection as permitted by that traffic signal (figure 1).

After the traffic signal had changed to a "Do Not Walk" indication in her direction of travel, the mishap pedestrian walked into the marked crosswalk on the far side of the intersection, walking from right to left from the point of view of the Cruise and Nissan vehicles.

The pedestrian crossed in front of the Cruise robotaxi first. During the 2.6 seconds that the pedestrian was in the crosswalk in front of the vehicle, the Cruise robotaxi accelerated from 5.5 mph to approximately 13.5 mph. This acceleration was due to a computer driver prediction that the pedestrian would be clear by the time the robotaxi entered the in-lane portion of the crosswalk.

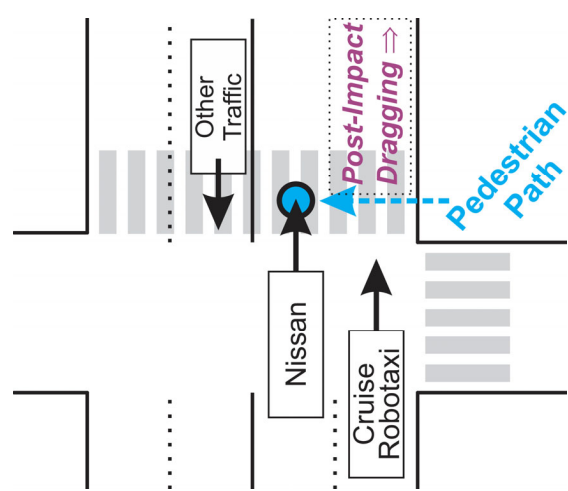

**Figure 1.** Simplified diagram of mishap; not to scale.

The pedestrian then started crossing the Nissan vehicle's lane, with that vehicle also continuing to accelerate. However, the pedestrian stopped mid-lane due to oncoming traffic in the opposing travel lanes. (Recall that the pedestrian started crossing after the lights had changed green in the vehicles' travel direction. This meant that there were moving vehicles in the farther lane of opposing-direction traffic that were blocking her progress.)

The pedestrian, apparently focusing on the oncoming traffic in the far lanes, did not seem to notice the oncoming Nissan. The Nissan did not slow down for her. At the time the Nissan struck her, she was walking at 2.6 mph in a direction redacted by the Cruise external report, but which might have been back toward her originating curb position on the far side of the robotaxi's travel lane.

The California Vehicle Code requires a vehicle to reduce speed when approaching a pedestrian in a crosswalk [5]. This requirement seems to apply regardless of traffic signal status. Both the Cruise robotaxi and the human-driven Nissan accelerated toward this pedestrian while she was in a crosswalk rather than decelerating as required by road rules.

The Cruise robotaxi's computer driver had been tracking both the Nissan and the pedestrian as other road users up to this point, including capturing a video frame of the impact. A post-crash reconstruction of the tracking data showed collision trajectories with the Nissan striking the pedestrian. However, the *Cruise computer driver "did not consider the potential of a* *collision between the Nissan and the pedestrian,"* so did not predict that event would change either the position or motion of the Nissan or the pedestrian. Apparently, the computer driver did not have a way to internally represent or predict the results of the impact between the Nissan and the pedestrian taking place in an adjacent lane.

The Nissan struck the pedestrian at 21.7 mph without ever having applied brakes. The human driver of the Nissan fled the scene, and as of this writing has not been apprehended.

After a 1.73 second long interaction with the Nissan, the pedestrian separated from the Nissan and entered the path of the Cruise robotaxi. The Cruise computer driver lost target track on the pedestrian during the impact with the Nissan. Intermittent detections ending at the last correct classification as a pedestrian continued until 0.3 seconds before impact by the Cruise robotaxi. Sustained pedestrian classification and trajectory tracking was not regained during the rest of the mishap sequence.

At 0.41 seconds before impact, the Cruise robotaxi detected an imminent collision with unclassified "occupied space" that the vehicle logic had been programmed to assume was a pedestrian, even though the object in that space was not considered to be reliably classified as a pedestrian. If the vehicle had begun braking immediately when the pedestrian entered its lane, there would have been just enough stopping distance available to completely avoid a collision. However, aggressive braking commands were not issued by the computer driver for about half a second after the pedestrian entered the lane 0.78 seconds before impact, resulting in a speed reduction due to braking of only 0.5 mph, with the robotaxi still traveling 18.6 mph at the time it impacted the pedestrian.

The pedestrian was first on the hood of the robotaxi, then later fell under the left front wheel. After impact, braking commands from the computer driver had time to take effect, resulting in aggressive post-impact braking. The pedestrian apparently remained largely under the vehicle and in the vicinity of the left front wheel until the robotaxi came to a stop 1.78 seconds after robotaxi impact.

**Post-crash dragging**

At this point, one would have expected the robotaxi to contact the CRAC to request guidance. Such a request was in fact initiated. However, the computer driver decided to attempt to move out of the travel lane without waiting for remote guidance.

Approximately 0.05 seconds after it had come to a stop, the Cruise robotaxi started a maneuver intended to pull it out of the travel lane, reaching a speed of 7.7 mph. This maneuver might have lasted for up to 100 feet of travel.

The robotaxi incorrectly diagnosed the pedestrian strike as having been a side impact rather than a frontal run-over impact, and was programmed in the case of a side impact to invoke the pull-over logic. Possibly this immediate pull-over behavior was programmed in response to extensive public and local governmental criticism for blocking travel lanes in previous, highly publicized non-injury incidents. The redacted external investigation report does not give a reason for this pull-over design strategy beyond a desire to achieve a so-called "minimal risk condition."

However, the pedestrian was still entrapped under the vehicle. This post-impact maneuver dragged her down the road until excessive wheel slip caused the vehicle to trigger a failure response with a gradual slowing to a stop. (The wheel slip was not recognized by the computer driver as being due to the pedestrian's body obstructing vehicle motion.) The total dragging distance was 20 feet.

The vehicle came to rest with its left rear tire on top of the pedestrian's legs, and her body under the rear of the vehicle. (See [6] for scene photos. Even in light of these photos, the general public did not understand that a dragging had taken place, instead attributing the scene to the pedestrian having been run over during a single impact event.)

Post-crash analysis revealed that a portion of the pedestrian's legs were visible in a robotaxi side-view camera throughout the event. However, the computer driver did not recognize that image as an entrapped pedestrian.

**Immediate mishap response**

An immediate support team response to the mishap was triggered by an automatic transmission of an initial crash video clip to the contractor-staffed CRAC in Arizona. This and a subsequent video clip showed the initial impact, but not the dragging portion of the sequence. However, due to the computer driver's immediate execution of the pullover maneuver, the dragging portion of the mishap sequence was already well underway before CRAC could possibly have intervened.

Approximately 2 minutes after the crash, a bystander called the public emergency services phone number ("911" in the US), with city first responders arriving on scene within 8 minutes. There is no indication that Cruise or its contractors ever notified emergency services of an injury collision with a pedestrian, although CRAC personnel clearly realized that such a collision had occurred when they reviewed the initial video from the vehicle.

Emergency responders asked CRAC via a vehicle-hosted voice link to immobilize the vehicle for safe extrication of the entrapped pedestrian. Different organizations of San Francisco-based Cruise support contractors arrived on-scene between 10 and 45 minutes after the crash. They had ample reason to believe a dragging of some sort had occurred based on post-crash robotaxi positioning past the crosswalk as well as a trail of victim blood and skin patches on the pavement.

Cruise staff, headquartered in San Francisco, were notified of a collision, initially classified as "minor" due to an apparent communication breakdown with CRAC and local response support contractors. (This and other contractor communication breakdowns are not explored in depth by the external investigation report.) The severity was later upgraded to "major with injury" two hours after the crash. These two severity determinations triggered escalating crisis response processes within Cruise.

The incident response playbook was said to be "too manually intensive," and Cruise quickly deviated from their planned crisis response procedures. In particular, Cruise fixated almost exclusively on media messaging and corporate image damage control, feeling that they were "under siege." The main messaging theme was attempting to focus attention on the Nissan human driver as having been the primary cause of the mishap.

External investigators claim a lack of "conclusive evidence" that Cruise employees (as opposed to contractors) knew of the pedestrian dragging portion of the mishap sequence until early the next morning. This created an apparent knowledge gap between Cruise and the city responders, who likely had inferred a dragging incident based on physical evidence at the crash scene, but might not have known about the initial Nissan impact. On the other hand, the Cruise staff knew about the Nissan impact, but are said not to have known a post-impact dragging event had taken place. Cruise leadership focused on attempting to counter any narrative that the robotaxi bore some blame for injuries to the pedestrian. Once adopted, this focus on avoiding blame on Cruise seemed to persist even after learning of the dragging aspect of the mishap.

**Regulatory and messaging responses**

By 3:45 AM San Francisco time the Cruise on-line "war room" team and other staff encompassing at least 100 employees were unambiguously told via a Slack internal messaging system that the pedestrian dragging portion of the mishap had occurred. The regulatory and broader media messaging efforts gained momentum after that time, although the messaging plan still downplayed the dragging portion of the mishap.

The messaging and outreach efforts the day after the crash had three parts: public media, state/local government, and federal government.

Media reporters were shown a version of the crash video that stopped before the pedestrian dragging portion of the mishap. Reporters were allowed to view the video via screen sharing, but were not given a copy. Some media reports conveyed a message that the Cruise vehicle was entirely responsible for the collision and the pedestrian's injuries. No mention of pedestrian dragging was made to any public media outlets because Cruise leadership believed they "had no obligation to share anything with the press." As the situation evolved, decisions were made to avoid additional disclosure of known facts out of fear of initiating a fresh negative news cycle.

State and local government representatives were shown a full length video, but no affirmative mention was made of the pedestrian dragging portion of the mishap. The strategy was said to be to let the video "speak for itself." Cruise did not affirmatively mention or call attention to the dragging to any external party during the post-crash discussions on the day after the crash.

City emergency responder representatives, who regularly deal with crash analysis involving emergency vehicles, noticed the dragging portion and asked questions. However, city officials have no regulatory power over robotaxi operations in California.

State and federal government officials, including CA DMV, apparently did not notice the dragging, or possibly did not see it due to Internet connectivity issues. The Cruise attitude was one of relief that the regulators did not ask about the dragging, which Cruise considered to be their biggest issue. Cruise staff felt they had "dodged a bullet" due to regulators not noticing the dragging.

Cruise decided not to do a safety stand-down after the mishap, and kept operating. Cruise leadership justified this decision by citing "overall driving and safety records" while minimizing the importance of the event as being an "extremely rare event" that was an "edge case."

CA DMV eventually realized that a dragging had occurred, demanding a copy of the full video and reviewing it starting with a meeting on October 13. On October 24, DMV suspended Cruise's California operating permits. Arguably the suspension was more a response to CA DMV feeling they had been duped into not understanding the pedestrian dragging had taken place rather than a reaction to the facts of the mishap itself.

At the federal level, Cruise submitted required 1-day and 10-day crash reports to the US National Highway Traffic Safety Administration (NHTSA) that omitted the pedestrian dragging portion of the mishap. The external investigation report claims a lack of administrative oversight caused this – along with providing written evidence showing that at least some explicit approval oversight of the report wording did occur. A required 30-day report to NHTSA submitted after the CA DMV permit suspension mentioned the pedestrian dragging.

In the aftermath of these events, many top Cruise leaders including the CEO were sacked, followed by a 24% workforce cut. Public road operations were suspended for months. Much closer oversight by the Cruise parent company, General Motors, is likely to be in place for the foreseeable future, with a long, slow path to resuming operations [7]. The injured pedestrian will reportedly receive between USD $8 and $12 million in settlement compensation [8].

**What went right**

Some things went well during the mishap sequence and following events based on engineering work and organizational preparations made before the crash.

... The Cruise robotaxi treated an unidentified object in its path as a potential pedestrian rather than a disregarded object, and executed a hard braking maneuver in response.

... Notification of an adverse event and a video of the initial crash of the robotaxi into the pedestrian were automatically sent to the CRAC remote response team within seconds of the pedestrian crash.

... Cruise employees formed crisis response teams and set up a virtual war room via a company-wide notification system in a timely manner.

... Cruise reached out to regulators and media the day after the crash in an orchestrated manner with management-approved messaging.

**What went wrong**

The list of things that went wrong both during and after the mishap is somewhat more extensive. We highlight important lessons for companies building and operating autonomous vehicles.

... The Cruise computer driver behaved in a way that was apparently inconsistent with California road rules by accelerating toward a pedestrian in a crosswalk. This provided it with less reaction time when the pedestrian re-entered its lane. Such behavior might be seen as negligent driving (inconsistent with road rules) that ultimately contributed to the initial harm the vehicle inflicted on a pedestrian. While the preceding human-driven Nissan collision with the pedestrian is certainly a consideration, the Cruise robotaxi might have avoided a collision entirely if it had slowed down while the pedestrian was in the crosswalk in its travel direction to give itself more reaction time as a defensive driving tactic.

... The Cruise computer driver lost track of the pedestrian during and after impact with the Nissan. Moreover, the computer driver did not have the ability to incorporate a readily predictable impact between those other road users only a few feet away into its world model. While Cruise staff considered the specific scenario involved in this mishap to be "unrealistic" and an "insane hypothetical" during the initial mishap analysis (per the external investigation report), pedestrians being hit and thrown into other lanes is quite foreseeable. Adding a capability to detect and mitigate risks due to nearby crashes should help reduce the risk of future harm from robotaxis as secondary participants in mishap scenarios.

... The Cruise computer driver took approximately half a second to issue a hard braking command from the time the pedestrian entered its lane. While an immediate response to this last-chance opportunity could have completely avoided the impact, reasonable readers might believe that some non-zero latency is necessary for a real-world implementation. However, a response time that was even a little faster would likely have reduced the harm done by that initial impact. This mishap is a data point in a larger continuing

- discussion about the tradeoffs between false alarm "phantom braking" events and the need to react quickly to real obstacles in the path of an automated vehicle.

- The Cruise computer driver essentially forgot that it had hit a pedestrian, and initiated follow-up movement almost immediately after its post-impact stop. The external investigators were quite clear that a human driver would not be expected to make this mistake, especially given the vehicle bumps and tire spin resulting from running over the pedestrian with the front tire during the initial impact. Some way of accounting for the potential presence of vulnerable road users who have been dropped from tracking should be considered to avoid a similar mishap in the future.

- The computer driver was apparently not trained on data showing pedestrian legs protruding from under the vehicle in its camera view. While some might say that avoiding the pedestrian dragging would require a missing under-vehicle sensor, in fact the pedestrian dragging could have been avoided for this mishap by recognizing pedestrian legs seen by the left side camera. This illustrates that rare events beyond normal driving conditions might be missing from training data.

- While there was significant public pressure to move any out-of-service robotaxi out of travel lanes quickly, programming the computer driver to move the vehicle after any impact that potentially involved a pedestrian was a problematic strategy. In this case, someone had made a decision that programming the computer driver to move the vehicle after a side impact was acceptable. However, in this mishap, due to ineffective tracking of the pedestrian, a frontal impact was incorrectly interpreted as a side impact by the system. Given the current immature state of the technology, it seems unwise for a computer driver to be permitted to move a vehicle after an injury mishap without human driver input or a very fast response by a remote operator.

- According to the narrative supplied by the external investigation, there were substantive communication failures between the remote contractor support team, the on-the-ground contractor support team, and the Cruise corporate team leading to delays in understanding the severity and circumstances of the mishap. However, the exact nature and causes of these failures were not examined in any substantive depth. Regardless of the details of this mishap, the role miscommunication is said to have played underscores the importance of having robust communication channels among different teams, locations, and organizations, especially in times of crisis.

- Neither the vehicle, CRAC, nor other Cruise-affiliated organization notified emergency services of a collision with a pedestrian. It seems unlikely that CRAC would have had any way of knowing that a bystander had already called emergency responders until responders arrived on scene, indicating that some sort of process failure resulted in the non-notification. Emergency services notification should be included in response plans and routinely exercised in practice drills.

- A decision not to be proactive in communicating the dragging portion of the mishap to regulators (and to some extent the public) arguably resulted in much worse outcomes for the company and management team. A fear of initiating a fresh negative news cycle prevented correcting overly favorable media stories. The cover-up – whether explicit or inadvertent – is always worse, regardless of intention.

- By not initiating a safety stand-down after a severe pedestrian injury, Cruise lost the opportunity to concentrate resources on an immediate identification and correction of both process and technical failures that contributed to the mishap. Additionally, Cruise missed an opportunity for a public demonstration of placing a priority on safety. A safety stand-down of some sort should be a reflexive response to a high-profile loss event.

It should be noted that there are redactions and unaddressed topics in the external report material used as a basis for this paper. For example, regulators were not interviewed by the authors of the external report, and contractor interview coverage was sparse. Also, there was no detailed analysis of CRAC logs and communication records. Many technical topics relevant to safety such as risk analysis, test strategy, backlog of unmitigated hazards, safety management system, and so on are out of scope for the external report and therefore not discussed herein.

Additionally, one ought to presume that a report written by a law firm paid by Cruise would frame the narrative surrounding presented facts and resolve any potential ambiguities in a way maximally favorable to Cruise's business interests.

Nonetheless, we believe that the basic facts are related accurately enough in the external investigation report that the observations and lessons to be learned are valid.

**Higher level lessons**

Some higher-level, more general lessons can be drawn from this case study.

Cruise persistently attempted to deflect blame by messaging that this mishap was (a) initiated by a human driver in the Nissan, (b) involved a jaywalking pedestrian who crossed against the traffic signal light, and (c) was a rare and in their opinion unforeseeable case. However, the human driver of the Nissan was not involved in initiating the pedestrian dragging portion of the mishap. Additionally, both vehicles failed to slow down for a pedestrian in a crosswalk as stated in California road rules.

A messaging strategy of blaming other road users for contributing to a mishap does not mean the harm didn't happen, and does not absolve the robotaxi designers and operators of their obligation to mitigate risks.

Practical safety on public roads tends to be dominated by comparatively infrequent mishaps. (For example, fatalities are commonly measured per 100 million miles or per billion kilometers.) Saying something is "rare" does not necessarily mean the overall risk is acceptable if the harm to a road user is severe – or even if the risk of reputational harm to the organization is too high.

While a pedestrian trapped under the robotaxi was likely not included in the computer driver's training data (either observed on roads or simulated), many rare high-consequence scenarios present unacceptable risk. And many of these risks are likely to involve off-nominal operational situations. Traditional safety engineering approaches such as hazard and risk analysis should be used to track such situations to resolution before they have a chance to happen on public roads. (We do not know for this particular mishap if risks related to the mishap scenario were tracked and accepted, or if they were not being tracked.)

It is important for autonomous system designers to keep in mind that their model of the external world will always have simplifications and omissions. These can be particularly problematic in an off-nominal or loss scenario. While it might be that a design team decides it is impractical to predict the trajectory of a nearby pedestrian after a collision with another vehicle, hazard analysis should dictate responding to that situation to mitigate risk somehow, such as slowing down to provide more response time rather than continuing to drive past such a high-probability impending nearby collision at normal speed.

Additionally, this mishap underscores the challenges of building an accurate world model of a post-crash scene, even with the numerous sophisticated sensors installed on high-end robotaxi vehicles. Pulling a vehicle to the side of the road is far from the all-purpose "minimal risk" strategy that is sometimes attributed to that maneuver. Significant thought must be given to unusual circumstances that might be present after a crash, as well as how a remote human operator might effectively gain situational awareness in an abnormal operational environment with potentially damaged vehicle sensors.

From an operational perspective, the importance of remote management teams has been getting less attention than it should. Procedures should not only be defined but also practiced regularly to ensure timely and effective responses to incidents. This includes robust mock exercises involving support contractors, operations staff, senior leadership, and corporate crisis response teams. Operational support teams will be required for the indefinite future to handle post-mishap responses, even as the need to routinely intervene in mundane driving decisions diminishes.

Robotaxi developers might rethink their strategy of deploying still-in-development technology without in-vehicle safety drivers. It seems likely that the pedestrian dragging would have been avoided entirely if an in-vehicle safety driver had been there to execute an emergency autonomy shutdown and then make a call to emergency services. Moreover, a large fraction of the other less severe robotaxi incidents that have occurred in San Francisco such as extended blocking of travel lanes and interfering with emergency responders could also have been avoided with in-vehicle backup drivers.

From an organizational perspective, the automotive industry should re-think its current default tactic of never disclosing any information unless that information is explicitly requested by a regulator. Lack of regulatory transparency and lack of proactive disclosure arguably hurt Cruise far more than running over a pedestrian and dragging her down the street.

Moreover, this lack of public transparency undoubtedly put pressure on regulators to react forcefully to the situation.

Future significant mishaps are inevitable, because public road travel has non-zero risk. The external investigation and subsequent public messaging from Cruise have emphasized improving regulatory interactions and regaining the trust of regulators. One hopes that Cruise is also addressing the numerous other issues we identify that go far beyond avoiding another botched regulatory interaction in the future. The organization should go far past controlling its messaging narrative and do everything reasonable to mitigate risk and ensure public road safety.

In the wake of this mishap, other injury crashes, and a continual stream of embarrassing press, the robotaxi industry faces a significant trust deficit with regulators, legislators, and the public. Improved transparent emphasis on safety is required to regain stakeholder trust. That transparency should start with every company explaining how it is addressing the lessons to be learned from this tragic mishap.

## ■ REFERENCES

**Philip Koopman** Prof. Philip Koopman is an internationally recognized expert on Autonomous Vehicle (AV) safety whose work in that area spans over 25 years. He is also actively involved with AV policy and standards as well as more general embedded system design and software quality. His pioneering research work includes software robustness testing and run time monitoring of autonomous systems to identify how they break and how to fix them. He has extensive experience in software safety and software quality across numerous transportation, industrial, and defense application domains including conventional automotive software and hardware systems. He originated the UL 4600 standard for autonomous system safety issued in 2020. He is a faculty member of the Carnegie Mellon University ECE department where he teaches software skills for mission-critical systems. In 2018 he was awarded the highly selective IEEE-SSIT Carl Barus Award for outstanding service in the public interest for his work in promoting automotive computer-based system safety. In 2022 he was named to the National Safety Council's Mobility Safety Advisory Group. In 2023 he was named the International System Safety Society's Educator of the Year. He is the author of the books: Understanding Checksums & Cyclic Redundancy Codes (2024), How Safe is Safe Enough: measuring and predicting autonomous vehicle safety (2022), The UL 4600 Guidebook (2022) and Better Embedded System Software (2010). Contact him at koopman@cmu.edu.